%% file: main.tex
\documentclass{article}

\usepackage[nonatbib, final]{neurips_2022}
\usepackage[numbers, sort&compress]{natbib}

\input{math_commands.tex}

\usepackage[utf8]{inputenc} % allow utf-8 input
\usepackage[T1]{fontenc}    % use 8-bit T1 fonts
\usepackage{hyperref}       % hyperlinks
\usepackage{url}            % simple URL typesetting
\usepackage{booktabs}       % professional-quality tables
\usepackage{amsfonts}       % blackboard math symbols
\usepackage{nicefrac}       % compact symbols for 1/2, etc.
\usepackage{microtype}      % microtypography
\usepackage{xcolor}         % colors
\usepackage{caption}
\usepackage{multirow}
\usepackage{graphicx}
\usepackage{amsmath}
\usepackage[title]{appendix}
\usepackage{wrapfig}
\usepackage[bottom]{footmisc}

\title{Homomorphic Self-Supervised Learning}

\author{T. Anderson Keller \\
Apple \\
\texttt{t.anderson.keller@gmail.com} \\
\And
Xavier Suau \\
Apple \\
\texttt{xsuaucuadros@apple.com} \\
\And
Luca Zappella \\
Apple \\
\texttt{lzappella@apple.com}
}

\begin{document}
\maketitle

\begin{abstract}
In this work, we observe that many existing self-supervised learning algorithms can be both unified and generalized when seen through the lens of equivariant representations. Specifically, we introduce a general framework we call \emph{Homomorphic Self-Supervised Learning}, and theoretically show how it may subsume the use of input-augmentations provided an augmentation-homomorphic feature extractor. We validate this theory experimentally for simple augmentations, demonstrate how the framework fails when representational structure is removed, and further empirically explore how the parameters of this framework relate to those of  traditional augmentation-based self-supervised learning. We conclude with a discussion of the potential benefits afforded by this new perspective on self-supervised learning.
\end{abstract}

\input{sections/introduction}

\input{sections/background}

\input{sections/latent_augmentations}
\input{sections/experiments}

\input{sections/discussion}



\bibliography{iclr2022_conference}
\bibliographystyle{iclr2022_conference}

\newpage
%%%%%%%%%%%%%%%%%%%%%%%%%%%%%%%%%%%%%%%%%%%%%%%%%%%%%%%%%%%%
\section*{Checklist}

%%% BEGIN INSTRUCTIONS %%%
The checklist follows the references.  Please
read the checklist guidelines carefully for information on how to answer these
questions.  For each question, change the default \answerTODO{} to \answerYes{},
\answerNo{}, or \answerNA{}.  You are strongly encouraged to include a {\bf
justification to your answer}, either by referencing the appropriate section of
your paper or providing a brief inline description.  For example:
\begin{itemize}
  \item Did you include the license to the code and datasets? \answerYes{See Section~\ref{gen_inst}.}
  \item Did you include the license to the code and datasets? \answerNo{The code and the data are proprietary.}
  \item Did you include the license to the code and datasets? \answerNA{}
\end{itemize}
Please do not modify the questions and only use the provided macros for your
answers.  Note that the Checklist section does not count towards the page
limit.  In your paper, please delete this instructions block and only keep the
Checklist section heading above along with the questions/answers below.
%%% END INSTRUCTIONS %%%

\begin{enumerate}

\item For all authors...
\begin{enumerate}
  \item Do the main claims made in the abstract and introduction accurately reflect the paper's contributions and scope?
    \answerYes{}
  \item Did you describe the limitations of your work?
    \answerYes{See Discussion Section $\ref{sec:discussion}$ and Appendix \ref{appendix:A}}
  \item Did you discuss any potential negative societal impacts of your work?
    \answerYes{See Appendix \ref{appendix:E}}
  \item Have you read the ethics review guidelines and ensured that your paper conforms to them?
    \answerYes{}
\end{enumerate}

\item If you are including theoretical results...
\begin{enumerate}
  \item Did you state the full set of assumptions of all theoretical results?
    \answerYes{See Section \ref{sec:s-ssl}}
        \item Did you include complete proofs of all theoretical results?
    \answerYes{See Section \ref{sec:s-ssl}}
\end{enumerate}

\item If you ran experiments...
\begin{enumerate}
  \item Did you include the code, data, and instructions needed to reproduce the main experimental results (either in the supplemental material or as a URL)?
    \answerNo{May be released at a later date}
  \item Did you specify all the training details (e.g., data splits, hyperparameters, how they were chosen)?
    \answerYes{See Appendix \ref{appendix:C}}
        \item Did you report error bars (e.g., with respect to the random seed after running experiments multiple times)?
    \answerYes{See Section \ref{sec:experiments}}
        \item Did you include the total amount of compute and the type of resources used (e.g., type of GPUs, internal cluster, or cloud provider)?
    \answerYes{See Section \ref{sec:experiments}}
\end{enumerate}

\item If you are using existing assets (e.g., code, data, models) or curating/releasing new assets...
\begin{enumerate}
  \item If your work uses existing assets, did you cite the creators?
    \answerYes{See Appendix \ref{appendix:C}}
  \item Did you mention the license of the assets?
    \answerYes{See Appendix \ref{appendix:C}}
  \item Did you include any new assets either in the supplemental material or as a URL?
    \answerNA{}
  \item Did you discuss whether and how consent was obtained from people whose data you're using/curating?
    \answerNA{}
  \item Did you discuss whether the data you are using/curating contains personally identifiable information or offensive content?
    \answerNA{}
\end{enumerate}

\item If you used crowdsourcing or conducted research with human subjects...
\begin{enumerate}
  \item Did you include the full text of instructions given to participants and screenshots, if applicable?
    \answerNA{}
  \item Did you describe any potential participant risks, with links to Institutional Review Board (IRB) approvals, if applicable?
    \answerNA{}
  \item Did you include the estimated hourly wage paid to participants and the total amount spent on participant compensation?
    \answerNA{}
\end{enumerate}

\end{enumerate}

%%%%%%%%%%%%%%%%%%%%%%%%%%%%%%%%%%%%%%%%%%%%%%%%%%%%%%%%%%%%
\newpage

\appendix
\begin{appendices}
\input{sections/appendix/appendix_A}

\input{sections/appendix/appendix_C}
\input{sections/appendix/appendix_D}
\input{sections/appendix/appendix_E}
\end{appendices}

\end{document}

%% file: math_commands.tex
\usepackage{amsmath,amsfonts,bm}

\def\eqref#1{equation~\ref{#1}}
\def\1{\bm{1}}

\def\vpsi{{\boldsymbol{\psi}}}

\def\va{{\bm{a}}}
\def\vb{{\bm{b}}}

\def\vg{{\bm{g}}}

\def\vx{{\bm{x}}}
\def\vy{{\bm{y}}}
\def\vz{{\bm{z}}}

\DeclareMathAlphabet{\mathsfit}{\encodingdefault}{\sfdefault}{m}{sl}
\SetMathAlphabet{\mathsfit}{bold}{\encodingdefault}{\sfdefault}{bx}{n}

\def\gG{{\mathcal{G}}}

\def\gL{{\mathcal{L}}}

\def\gX{{\mathcal{X}}}
\def\gY{{\mathcal{Y}}}
\def\gZ{{\mathcal{Z}}}

\def\sR{{\mathbb{R}}}

\def\sZ{{\mathbb{Z}}}

\newcommand{\E}{\mathbb{E}}

%% file: sections/introduction.tex
\section{Introduction}
\label{sec:introduction}

\looseness=-1
Many self-supervised learning (SSL) techniques can be colloquially defined as representation learning algorithms which extract approximate supervision signals directly from the input data itself \citep{ssl-dark-matter}. In practice, this supervision signal is often obtained by performing symmetry transformations of the input with respect to task-relevant information, meaning the transformations leave task-relevant information unchanged, while altering task-irrelevant information. Numerous theoretical and empirical works have shown that by combining such symmetry transformations with contrastive objectives, powerful lower dimensional representations can be learned which support linear-separability, identifiability of generative factors, and reduced sample complexity \citep{SSL_MV, marco_MV, wang2022chaos, ji2021power, ssl_content_style, approx_CI, ddn, theory_cl, cl_mv, cmc, byol, simclr, simsiam}. One rapidly developing domain of deep learning research which is specifically focused on the structured and accurate representation of the input with respect to symmetry transformations is that of equivariant neural networks \citep{gcnn,  cohen2016steerable, worrall2017harmonic, weiler20183d, scalespaces, finzi2020generalizing, finzi2021emlp}.
In this work, we study the properties of SSL algorithms when equivariant neural networks are used as backbone feature extractors. Interestingly, we find a convergence of existing loss functions from the literature, and ultimately generalize these with the framework of \emph{Homomorphic Self-Supervised Learning}. 
\begin{figure}[h]
\caption{Overview of Homomorphic-SSL (left) and its relation to traditional Augmentation-based SSL (right). Positive pairs extracted from the lifted dimension ($\theta$) of a rotation equivariant network (G-conv) are equivalent to pairs extracted from the separate representations of two rotated images.}
\centering
\vspace{-2mm}
\includegraphics[width=0.90\textwidth]{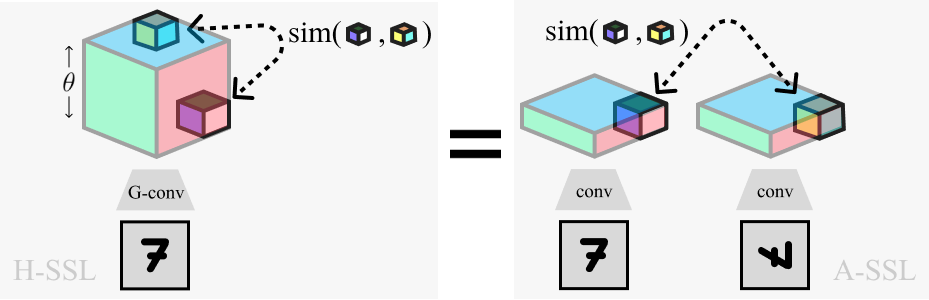}
\label{fig:equiv}
\vspace{-6mm}
\end{figure}
\raggedbottom

%% file: sections/background.tex
\section{Background}
\vspace{-1mm}
\label{sec:background}

\paragraph{Equivariance} 
The map $f: \gX  \rightarrow \gZ$ is said to be equivariant with respect to the group $\gG = (G, \cdot)$ if
\begin{equation}
\label{eqn:equivariance}
  \exists \Gamma_g \ \ \ \text{such that} \ \ \ \  f(T_g [\vx]) = \Gamma_g [f(\vx)] \ \ \ \ \forall g \in G\ ,
\end{equation}
where $G$ is the set of all group elements, $\cdot$ is the group operation, $T_g$ is the representation of the transformation $g \in G$ in input space $\gX$, and $\Gamma_g$ is the representation of the same transformation in output space $\gZ$. If $T_g$ and $\Gamma_g$ are formal group representations \cite{serre} such maps $f$ are termed group-homomorphisms since they can be seen to preserve the structure of the group in the output space.
There are many different methods for constructing group equivariant neural networks, resulting in different representations of the transformation in feature space $\Gamma_g$. In this work, we consider only discrete groups $\gG$ and networks which admit regular representations for $\Gamma$ (see Appendix \ref{appendix:C} for an example). Specifically, we denote the output of our network $f(\vx) = \vz \in \sR^{C \times |G|}$, where $C$ is the number of output channels. As a simple example, a standard convolutional layer would have all height ($H$) and width ($W$) spatial coordinates as the set $G$, giving $\vz \in \sR^{C \times HW}$. A group-equivariant neural network \cite{gcnn} which is equivariant with respect to the the group of all integer translations and 90-degree rotations ($p4$) would thus have a feature multiplicity four times larger ($\vz \in \sR^{C \times 4HW}$), since each spatial element is associated with the four distinct rotation elements $(0^o, 90^o, 180^o, 270^o)$. Such a rotation equivariant network is depicted in Figure \ref{fig:equiv} with the `lifted' rotation dimension extended along the vertical axis ($\theta$). In both the translation and rotation cases, the regular representation $\Gamma_g$ acts by permuting the representation along the group dimension, leaving the feature channels unchanged.

% \vspace{-1mm}
\paragraph{Notation} The vector of features (channels) at a specific group element $g$ is sometimes called a `fiber' \cite{cohen2016steerable}. In this work we use the following shorthand for indexing fibers according to their group element $g$: $\vz(g) \equiv \vz_{:,g} \in \sR^{C}$. Similarly, the set of fibers corresponding to an ordered set of group elements $\vg$ can be called a `fiber bundle' which we denote: $\vz(\vg) = [\vz(g)\  |\  g \in \vg] \in \sR^{|\vg|C}$. Fiber bundles can be seen in Figure \ref{fig:equiv} as the small cubes being compared (with the fibers themselves extending into the undepicted fourth dimension). Using this notation, we can define the action of $\Gamma_g$ as: $\Gamma_g [\vz(\vg_0)] = \vz(g^{-1} \cdot \vg_0)$. Thus $\Gamma_g$ can be seen to move the fibers from `base' locations $\vg_0$ to a new ordered set of locations $g^{-1} \cdot\vg_0$. We highlight that order is critical for our definition since a transformation such as rotation may simply permute $\vg_0$ while leaving the unordered set intact.

\paragraph{Augmentation-based SSL (A-SSL)} 

%% file: sections/latent_augmentations.tex
Many state of the art SSL approaches rely on input augmentations in order to selectively extract task-relevant information. One prominent framework, SimCLR \citep{simclr}, trains a backbone feature extractor $f(\cdot)$ to minimize a contrastive loss applied to the representations of two augmented versions of an image. Specifically, given a batch of $N$ input images $\mathbf{X} = \{\vx_1, \vx_2, \ldots, \vx_N\}$, a similarity function $\mathrm{sim}(\va, \vb) = \frac{\va^T \vb}{||\va||\cdot ||\vb||}$, and a non-linear `projection head' $h: \gZ \rightarrow \gY$, the SimCLR loss is given as:
\vspace{-1mm}
\begin{equation}
\label{eqn:simclr}
    \footnotesize
    \mathcal{L}_{\text{A-SSL}}(\mathbf{X}) = - \frac{1}{N} \sum_{i}^N \E_{g_1, g_2 \sim G} \log{\frac{\mathrm{exp}\Big(\mathrm{sim}\big(h(f(T_{g_1} [\vx_i])), h(f(T_{g_2}[\vx_i]))\big) / \tau \Big)}{\sum_{k \neq i}^{N} \sum^2_{j,l}  \mathrm{exp}\Big(\mathrm{sim}\big(h(f(T_{g_j} [\vx_i])), h(f(T_{g_l} [\vx_k]))\big) / \tau\Big)}} \ ,
\end{equation}
\normalsize
where $G$ is the set of all augmentations, $T_g[\vx]$ denotes the action of the sampled augmentation $g$ on the input, and $\tau$ is the `temperature' of the softmax. In this work, we will focus on the SimCLR objective for simplicity, but our analysis also applies to non-contrastive frameworks such as BYOL \cite{byol} and SimSiam \cite{simsiam}, provided the backbone is equivariant (i.e. augmentation-homomorphic). 

\section{Homomorphic Self-Supervised Learning}
\label{sec:s-ssl}
In this section we introduce Homomorphic Self-Supervised Learning (H-SSL) as a general framework for SSL with homomorphic encoders, and further show how many existing SSL algorithms can be both unified and generalized. To begin, consider an A-SSL objective such as Equation \ref{eqn:simclr} when $f$ is equivariant with respect to the input augmentation. By the definition of equivariant maps in Equation \ref{eqn:equivariance}, the augmentation commutes with the feature extractor: $h(f(T_g [\vx])) = h(\Gamma_g [f(\vx)])$. 
Thus, replacing $f(\vx_i)$ with its output $\vz_i = \vz(\vg_{0})$, and applying the definition of the operator, we get: 
\vspace{0mm}
\begin{equation}
    \footnotesize
    \label{eq:s-ssl}
     \mathcal{L}_{\text{H-SSL}}(\mathbf{X}) = - \frac{1}{N} \sum_{i}^N \E_{g_1, g_2 \sim G} \log{\frac{\mathrm{exp}\Big(\mathrm{sim}\big(h(  \vz_i(g_1^{-1} \cdot\vg_{0})), h(\vz_i(g_2^{-1} \cdot \vg_{0}))\big) / \tau\Big)}{\sum_{k \neq i}^{N} \sum_{j,l}^2  \mathrm{exp}\Big({\mathrm{sim}\big(h( \vz_i(g_j^{-1} \cdot \vg_{0})), h( \vz_k(g_k^{-1} \cdot \vg_{0}))}\big) / \tau\Big)}} \ .
\end{equation}
\normalsize
\looseness=-1
Ultimately, we see that $\gL_{\text{H-SSL}}$ subsumes the use of input augmentations by defining the `positive pairs' in the numerator as two fiber bundles from \emph{the same representation $\vz_i$}, simply indexed using two differently transformed base spaces $g_1^{-1} \cdot\vg_{0}$ and $g_2^{-1} \cdot \vg_{0}$ (depicted in Figure \ref{fig:equiv}). Interestingly, this loss highlights the base space $\vg_0$ as a parameter choice previously unexplored in the A-SSL frameworks.

A second interesting consequence of this derivation is the striking similarity of the $\gL_{\text{H-SSL}}$ objective and other existing SSL objectives which operate without explicit input augmentations to generate multiple views. Specifically, when applied to images, Greedy InfoMax (GIM) \citep{gim} and Contrastive Predictive Coding (CPC) \citep{cpc}\footnote{In CPC, the authors use an autoregressive encoder to encode one element of the positive pairs. In GIM, they find that in the visual domain, this autoregressive encoder is not necessary, and thus the loss reduces to a standard contrastive loss between the representations from raw spatial patches, as defined here.} use a similar InfoNCE-inspired loss (as in SimCLR) but between different spatial locations of a convolutional filter stack for a single image. Similarly, the `local' term in the Deep InfoMax objective (DIM(L)) \citep{dim} operates entirely within convolutional feature maps. Consequently, these losses are contained in our framework where $\gG$ is set to the 2D translation group, and $\vg_0$ is a small subset of the spatial coordinates. Since $\gL_{\text{H-SSL}}$ is also derived directly from $\gL_{\text{A-SSL}}$ (when $f$ is equivariant), we see that it provides a means to unify these previously distinct sets of SSL objectives. In Section \ref{sec:experiments} we validate this theoretical equivalence empirically. Furthermore, since $\gL_{\text{H-SSL}}$ is defined for transformation groups beyond translation, it can be seen to generalize these augmentation-free objectives in a way that we have not previously seen exploited in the literature. In Section \ref{sec:experiments} we include a preliminary exploration this generalization to scale and rotation groups.

%% file: sections/experiments.tex
\section{Experiments}
\label{sec:experiments}
We now empirically validate the derived equivalence of A-SSL and H-SSL in practice, and further reinforce our stated assumptions by demonstrating how H-SSL objectives are ineffective when representational structure is removed. We study how the parameters of H-SSL (topographic distance) relate to those traditionally used in A-SSL (augmentation strength), and finally explore how new parameter generalizations afforded by our framework (such as choices of $\vg_0$ and $\gG$) impact performance. 

\paragraph{Empirical Validation} For perfectly equivariant networks $f$, and sets of transformations which exactly satisfy the group axioms, the equivalence between Equations \ref{eqn:simclr} and \ref{eq:s-ssl} is exact. However, in practice, due to discretization, boundary effects, and sampling artifacts, even for simple transformations such as translation, equivariance has been shown to not be strictly satisfied \citep{cnnequi}. In Table \ref{tab:equivalence} we empirically validate our proposed theoretical equivalence between $\gL_{\text{A-SSL}}$ and $\gL_{\text{H-SSL}}$, showing a tight correspondence between the downstream accuracy of linear classifiers trained on representations learned via the two frameworks. Precisely, for each transformation (Rotation, Translation, Scale), we use a backbone network which is equivariant specifically with respect to that transformation (e.g. rotation equivariant CNNs, regular CNNs, and Scale Equivariant Steerable Networks (SESN) \cite{SESN}). In all settings, we take $\vz$ from the final possible layer and set $\vg_0$ to be a single fiber of dimension 128.

\begin{table}[h]
\vspace{-2mm}
\caption{
\looseness=-1
MNIST \cite{mnist}, CIFAR10 \cite{c10} and Tiny ImageNet \cite{tin} top-1 test accuracy (mean $\pm$ std. over 3 runs) of a detached classifier trained on the representations from SSL methods with different backbones. We compare $\gL_{\text{A-SSL}}$ and  $\gL_{\text{H-SSL}}$ with random frozen and fully supervised backbones. We see equivalence between A-SSL and H-SSL as desired from the first two columns, and often a significant improvement in performance for H-SSL methods when moving from Translation to generalized groups such as Scale. Full experiment details can be found in Appendix \ref{appendix:C}.}
\label{tab:equivalence}
\begin{center}
\small
\begin{tabular}{c c c c c c c} 
\toprule
  Dataset & Transformation & Backbone & A-SSL & H-SSL & Frozen & Supervised \\ 
 \cmidrule(lr){1-1} \cmidrule(lr){2-3} \cmidrule(lr){4-7}
 \multirow{3}{*}{MNIST}
  & Rotation  & Rot-Eq. & {68.2 $\pm$ 2.5} &  {70.3 $\pm$ 5.4} & \textit{87.2 $\pm$ 0.8} & \textit{99.4 $\pm$ 0.1} \\
 & Translation & CNN & {95.9 $\pm$ 0.3} & {96.0 $\pm$ 1.3} & \textit{94.1 $\pm$ 0.3} & \textit{99.2 $\pm$ 0.1} \\
 & Scale & SESN & {98.6 $\pm$ 0.1} & {98.3 $\pm$ 0.2} & \textit{94.7 $\pm$ 0.6} & \textit{99.3 $\pm$ 0.1}\\ 
 \cmidrule(lr){1-1} \cmidrule(lr){2-3} \cmidrule(lr){4-7}
 \multirow{3}{*}{CIFAR10}
  & Rotation & Rot-Eq. & {46.1 $\pm$ 0.6} & {48.3 $\pm$ 0.5} & \textit{38.4 $\pm$ 0.1} & \textit{73.0 $\pm$ 1.1}  \\
  & Translation & CNN & {39.2 $\pm$ 0.5} & {36.3 $\pm$ 1.1} & \textit{40.4 $\pm$ 0.2} & \textit{76.2 $\pm$ 1.4} \\
 & Scale & SESN & {59.4 $\pm$ 0.2} & {56.7 $\pm$ 0.4} & \textit{41.1 $\pm$ 0.6} & \textit{78.0 $\pm$ 0.2} \\ 
 \cmidrule(lr){1-1} \cmidrule(lr){2-3} \cmidrule(lr){4-7}
 \multirow{2}{*}{Tiny ImageNet}
  & Rotation & Rot-Eq. & {14.9 $\pm$ 0.3} & {13.5 $\pm$ 0.5} & \textit{6.1 $\pm$ 0.2} & \textit{22.5 $\pm$ 0.1}  \\
 & Scale & SESN & {16.2 $\pm$ 0.4} & {14.0 $\pm$ 1.3} & \textit{6.4 $\pm$ 0.2} & \textit{23.7 $\pm$ 0.2} \\ 
\bottomrule
 \end{tabular}
\end{center}
\vspace{-3mm}
 \end{table}

\paragraph{H-SSL Without Structure} To validate our assertion that $\gL_{\text{H-SSL}}$ requires a homomorphism, in Table \ref{tab:non_equivariant} we show the same models from Table \ref{tab:equivalence} without equivariant backbones. We observe  $\gL_{\text{H-SSL}}$ models perform significantly below their input-augmentation counterparts, and similarly to a `frozen' randomly initialized backbone baseline -- indicating the learning algorithm is no longer effective.

\vspace{-2mm}
\begin{table}[h]
\caption{An extension of Table \ref{tab:equivalence} with non-equivariant backbones. 
We see that the H-SSL methods perform similar to, or worse than, the frozen baseline when equivariance is removed, as expected.}
\label{tab:non_equivariant}
\begin{center}
\small
\begin{tabular}{c c c c c c c} 
\toprule
  Dataset & Transformation & Backbone & A-SSL & H-SSL & Frozen & Supervised \\ 
  \cmidrule(lr){1-1} \cmidrule(lr){2-3} \cmidrule(lr){4-7}
 \multirow{2}{*}{MNIST}
 & Translation & MLP &  87.6 $\pm$ 0.2 & 58.2 $\pm$ 0.5  & 83.0 $\pm$ 0.8 & 98.6 $\pm$ 0.1 \\
 & Scale & CNN ($6 \times CHW$) & 95.2 $\pm$ 0.1 & 87.2 $\pm$ 2.4 & 87.2 $\pm$ 0.6 & 99.3 $\pm$ 0.1\\ 
 \cmidrule(lr){1-1} \cmidrule(lr){2-3} \cmidrule(lr){4-7}
 \multirow{1}{*}{CIFAR10}
 & Scale & CNN ($6 \times CHW$) & 53.6 $\pm$ 0.2 & 37.5 $\pm$ 0.1 & 43.6 $\pm$ 0.3 & 67.9 $\pm$ 2.1 \\ 
\bottomrule
 \end{tabular}
\end{center}
 \end{table}
\vspace{-3mm}

\begin{figure}[h]
\caption{Study of the impact of new H-SSL parameters top-1 test accuracy. (Left) Test accuracy marginally increases as we increase total base space size $
\vg_0$. (Right) Test accuracy is constant or decreases as we increase the maximum distance between fiber bundles considered positive pairs.}
\vspace{-2mm}
\centering
\includegraphics[width=1.0\textwidth]{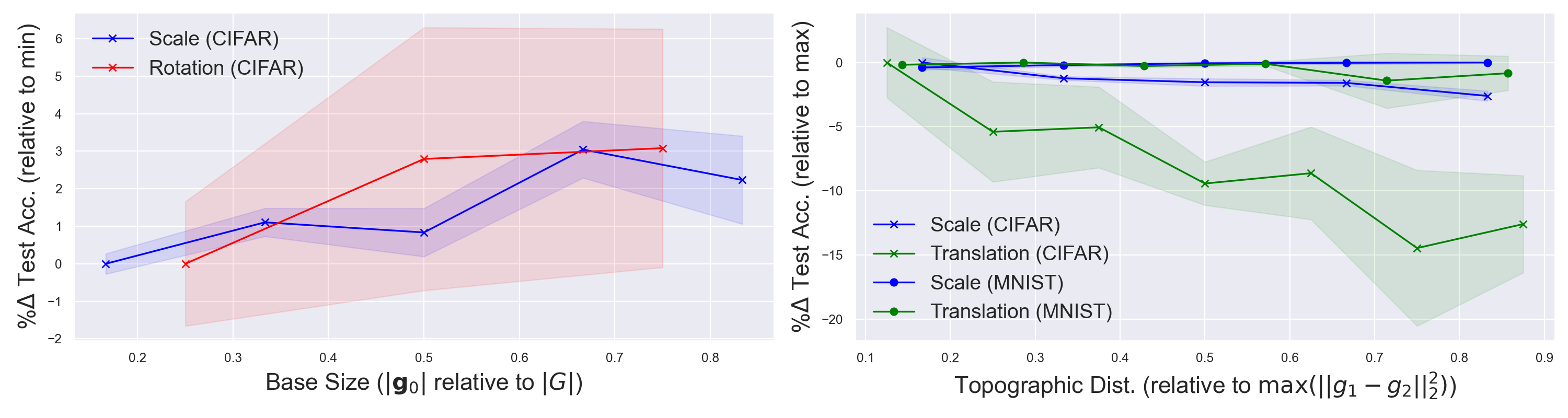}
\label{fig:size_topo_dist}
\end{figure}

\paragraph{Parameters of H-SSL} As discussed, The H-SSL framework highlights new parameter choices such as the base space $\vg_0$. In Figure \ref{fig:size_topo_dist} (left) we plot the $\%$-change in top-1 accuracy on CIFAR-10 as we increase the total size of $\vg_0$ from 1 (akin to DIM(L) losses) to $|G|-1$ (akin to SimCLR). We see a minor increase in performance as we increase the size, but note relative stability, again suggesting greater unity between A-SSL and H-SSL. In Figure \ref{fig:size_topo_dist} (right), we explore how the traditional notion of augmentation `strength' can be equated with the `topographic distance' between $g_1$ and $g_2$ and their associated fiber bundles. Here we approximate topographic distance as euclidean distance between group elements for simplicity ($||g_1 - g_2||_2^2$), where a more correct measure would be computed using the topology of the group. We see, in alignment with prior work \cite{what_makes_good_views}, that the strength of augmentation (and specifically translation distance) is an important parameter for effective self supervised learning, likely relating to the mutual information between fibers as a function of distance.

\vspace{-10mm}

%% file: sections/discussion.tex
\section{Discussion}
\label{sec:discussion}
\looseness=-1
In this work we have studied the impact of combining augmentation-homomorphic feature extractors with augmentation-based SSL objectives. In doing so, we have introduced a new framework which we call Homomorphic-SSL which illustrates an equivalence between previously distinct SSL methods when the homomorphism constraint is satisfied. Since it is not currently known how to construct neural networks which are analytically equivariant with respect to all input augmentations used in modern SSL, this constraint is precisely the greatest current limitation of this framework, and we expand on this limitation in Appendix \ref{appendix:A}. We therefore propose this work not as an improvment to the state of the art, but rather as a new perspective on SSL which provides a bridge to previously distant literature. Specifically, one field of research which appears particularly promising for future work is the integration of learned homomorphisms \cite{keller2021topographic, keurti2022homomorphism, connor2021variational, lconv, nptn} with H-SSL. In the H-SSL framework, a learned homomorphism can be seen as equivalent to a learned augmentation, providing a potential new avenue for approaching the extremely challenging \cite{pmlr-v163-blaas22a} but fruitful \cite{shi2022adversarial} goal of learned image augmentations. 

We additionally present this work as an attempt to renew interest in SSL objectives which operate without multiple inferences of a transformed image, such as Deep InfoMax \cite{dim} and Greedy InfoMax \cite{gim}, by allowing them to exploit the theoretical foundations developed for multi-view SSL \cite{cl_mv, ddn, SSL_MV, ssl_content_style, marco_MV, theory_cl}. Although DIM-like methods have to-date not yielded the same performance as their A-SSL counterparts, we believe the coupling between objective and network architecture is likely to yield more parallelizable algorithms which are therefore more scalable and biologically plausible \cite{gim}.

%% file: sections/appendix/appendix_A.tex
\section{Limitations}
\label{appendix:A}
Despite the demonstrated unification of existing methods, and benefits from generalization, we note that this approach is still significantly limited. Specifically, the equivalence between $\gL_{\text{A-SSL}}$ and $\gL_{\text{H-SSL}}$, and benefits afforded by this equivalence,  can only be realized if it is possible to analytically construct a neural network which is equivariant with respect to transformations of interest. Although the field of equivariant deep learning has made significant progress in recent years, state of the art techniques are still restricted to E($n$) and continuous compact and connected Lie Groups \cite{finzi2020generalizing, finzi2021emlp, cesa2022a, e2cnn}.
We believe in this regard, our analysis sheds some light on the success of methods which perform data augmentation over those which operate directly in feature space in recent literature -- it is simply too challenging with current methods to construct models with structured representations for the diversity of transformations needed to induce a sufficient set of invariances for linear separability of classes. 

In light of this, we believe that our framework specifically suggests a novel path forward via learned homomorphisms, \cite{keller2021topographic, keurti2022homomorphism, connor2021variational, lconv, nptn}, as mentioned in the Discussion Section \ref{sec:discussion}. Specifically, if new methods can learn symmetries from the data unsupervised, or minimally supervised, our framework yields a natural avenue by which existing SSL advances can be extended to feature-space methods. As a potential interim solution,  recent work \cite{amdim} has additionally shown that it is possible to train `hybrid' models that leverage both A-SSL and H-SSL objectives simultaneously, suggesting that this limitation may be circumvented by applying H-SSL losses only where beneficial, and otherwise supplementing with input space augmentation.

%% file: sections/appendix/appendix_C.tex
\section{Experiment Details}
\label{appendix:C}

\paragraph{Model Architectures} All models presented in this paper were built using the convolutional layers from the SESN \cite{SESN} library for consistency and comparability (\url{https://github.com/ISosnovik/sesn}). For scale equivariant models, we used the set of 6 scales $[1.0, 1.25, 1.33, 1.5, 1.66, 1.75]$. To construct the rotation equivariant backbones, we use only a single scale of $[1.0]$ and augment the basis set with four 90-degree rotated copies of the basis functions at $[0^o, 90^o, 180^o, 270^o]$. These rotated copies thus defined the group dimension. This technique of basis or filter-augmentation for implementing equivariance is known from prior work and has been shown to be equivalent to other methods of constructing group-equivariant neural networks \cite{filtra}. For translation models, we perform no basis-augmentation, and again define the set of scales used in the basis to a single scale $[1.0]$, thereby leaving only the spatial coordinates of the final feature maps to define the output group. 

On MNIST \cite{mnist}, we used a backbone network $f$ composed of three SESN convolutional layers with \# channels (32, 64, 128), kernel sizes (11, 7, 7), effective sizes (11, 3, 3), strides (1, 2, 2), padding (5, 3, 3), no biases, basis type `A', BatchNorm layers after each convolution, and ReLU activations after each BatchNorm. The output of this final ReLU was then considered our $\vz$ for contrastive learning (with $\gL_{A-SSL}$ and $\gL_{H-SSL}$) and was of shape $(128, S \times R, 8, 8)$ where $S$ was the number of scales for the experiment (either 1 or 6), and $R$ was the number of rotation angles (either 1 or 4). For experiments where the transformation studied was not translation, we average pool over the spatial dimensions before applying the projection head $h$ to achieve a consistent dimensionality of $128$. For classification, an additional SESN convolutional layer was placed on top with kernel size 7, effective size 3, stride 2, and no padding, thereby reducing the spatial dimensions to 1, and the total dimensionality of the input to the final linear classifier to 128.

On CIFAR10 we used a ResNet20 model composed of an initial SESN lifting layer with kernel size 7, effective size 7, stride 1, padding 3, no bias, basis type `A', and 9 output channels. This lifted representation was then processed by a following SESN convolutional layer of kernel size 7, effective size 3, stride 1, padding 3, no bias, basis type `A', and 64 output channels. This initial layer was followed by a BatchNorm and ReLU before being processed by three ResNet blocks of output sizes (128, 256, 512) and initial strides of (1, 2, 2). Each ResNet block is composed of 3 SESN Basic blocks as defined here (\url{https://github.com/ISosnovik/sesn/blob/master/models/stl_ses.py#L19}). The output of the third ResNet block was taken as our $\vz$ for contrastive learning (again for $\gL_{A-SSL}$ and $\gL_{H-SSL}$) of shape $(512, S \times R, 7, 7)$. Again, as for MNIST, for experiments where the transformation studied was not translation, we average pool over the spatial dimensions before applying the projection head $h$ to achieve a consistent dimensionality of $512$. For classification, the vector $\vz$ was first max-pooled along the scale/rotation group-axis ($S \times R$), followed by a BatchNorm, a ReLU, and average pooling over the remaining $7 \times 7$ spatial dimensions. Finally, we apply BatchNorm to this 512-dimensional vector before applying the non-linear projection head $h$.

On Tiny ImageNet we use a Resnet20 model which has virtually the same structure as the CIFAR10 model, but instead uses 4 ResNet blocks of output sizes (64, 128, 256, 512) and strides (1, 2, 2, 2). Furthermore, each ResNet block is composed of only 2 BasicBlocks for TIN instead of 3 for CIFAR10. Overall this results in a $\vz$ of shape $(512, S \times R, 4, 4)$, and a final vector for classification of size 512. We note that we do not include Translation results in Table \ref{tab:equivalence} for Tiny ImageNet precisely because the spatial dimensions of the feature map with this architecture are too small to allow for effective H-SSL training in the settings we used for other methods. 

All models used a detached linear classifier for computing the reported downstream classification accuracies, while the Supervised baselines used an attached linear layer (implying gradients with respect to the classification loss back-propagated though the whole network). All models additionally used an attached non-linear projection head $h$ constructed as an MLP with three linear layers. For MNIST these layers have of output sizes (128, 128, 128), while for CIFAR10 and TIN they have sizes (512, 2048, 512). There is a BatchNorm after each layer, and ReLU activations between the middle layers (not at the last layer). 

\paragraph{Training Details} For training we use the LARS optimizer with an initial learning rate of 0.1, and a batch size of 4096 for all models. We use an NCE temperature ($\tau$) of 0.1, half-precision training, a learning rate warm-up of 10 epochs, a cosine lr-update schedule, and weight decay of $1\times10^{-4}$. On MNIST we train for 500 epochs and on CIFAR10 and Tiny ImagNet (TIN) we train for 1300 epochs. On average each MNIST run took 1 hour to complete distributed across 8 GPUs, and each CIFAR10/TIN run took 10 hours to complete distributed across 64 GPUs. In total this amounts to roughly 85,000 GPU hours.

\paragraph{Empirical Validation} For the experiments in Table \ref{tab:equivalence}, we use two different methods for data augmentation, and similarly two different methods for selecting the representations ultimately fed to the contrastive loss for the A-SSL and H-SSL settings.

For A-SSL we augment the input at the pixel level by: randomly translating the image by up to $\pm$ $20\%$ of its height/width (for translation), randomly rotating the image by one of ($0^o, 90^o, 180^o, 270^o$) (for rotation), or randomly downscaling the image between $0.57$ and $1.0$ of its original scale. For S-SSL we use no input augmentations.

For both methods we use only a single fiber, meaning the base size $|\vg_0|$ is 1. For A-SSL, we randomly select the location $\vg_0$ for each example, but we use the same $\vg_0$ between both branches. For example, in translation, we compare the feature vectors for two translated images \emph{at the same pixel location}. Similarly, for scale and rotation, we pick a single scale or rotation element to compare for both branches. For H-SSL, we randomly select the location $\vg$ independently for each example \emph{and independently for each branch}, effectively mimicing the latent operator. 

\paragraph{H-SSL Without Structure}
In Table \ref{tab:non_equivariant}, we use the same overall model architectures defined above (3-layer model or ResNet20), but replace the individual layers with non-equivariant counterparts. Specifically, for the MLP, we replace the convolutional layers with fully connected layers with outputs (784, 1024, 2048). For the convolutional models (denoted CNN ($6 \times CHW$)), we replace the SESN kernels with fully-parameterized, non-equivariant counterparts, otherwise keeping the output dimensionality the same (resulting in the 6 $\times$ larger output dimension).

Furthermore, for these un-structured representations, in the H-SSL setting, we `emulate' a group dimension to sample `fibers' from. Specifically, for the MLP we simply reshape the 2048 dimensional output to ($16$,$128$), and select one of the 16 rows at each iterations. For the CNN, we similarly use the 6 times larger feature space to sample $\frac{1}{6}^{th}$ of the  elements as if they were scale-equivariant.

\paragraph{Parameters of H-SSL}
For Figure $\ref{fig:size_topo_dist}$ (left), we select patches of sizes from $1$ to $|G| - 1$ with no padding. In each setting, we similarly increase the dimensionality of the input layer for the non-linear projection head $h$ to match the multiplicative increase in the dimension of the input representation $\vz(\vg)$. For the topographic distance experiments (right), we keep a fixed base size of $|\vg_0| = 1$ and instead vary the maximum allowed distance between randomly sampled pairs $g_1$ \& $g_2$. 

%% file: sections/appendix/appendix_D.tex
\section{Extended Background}

\paragraph{Related Work}
Our work is undoubtedly built upon the the large literature base from the fields equivariant deep learning and self-supervised learning as outlined in Sections \ref{sec:introduction} and \ref{sec:background}. Beyond this background, our work is highly related in motivation to a number of studies specifically related to equivariance in self-supervised learning. Most prior work, however, has focused on the undesired invariances learned by A-SSL methods \cite{xiao2021what, tsai2022conditional} and on developing methods by which to avoid this through learned approximate equivariance \cite{dangovski2022equivariant, wang2021residual}. Our work is, to the best of our knowledge, the first to suggest and validate that the primary reason for the success of feature-space SSL objectives such as DIM(L) \cite{dim} and GIM \cite{gim} is due to their exploitation of equivariant backbones.

\paragraph{Group-Convolutional Neural Networks}
As discussed in Section \ref{sec:background}, we assume that the backbones used in this work are equivariant with respect to input augmentations, and further that they admit regular representations of those transformations in feature space. In this section we detail how such group-equivariant convolutional neural networks may be constructed via the group-convolution \cite{gcnn}: For a discrete group $\gG$, we denote the pre-activation output of a $\gG$-equivariant convolutional layer $l$ as $\vz^l$, with a corresponding input $\vy^l$. In practice these values are stored in finite arrays with a feature multiplicity equal to the order of the group in each space. Explicitly, $\vz^l \in \sR^{C_{out} \times |G_{out}|}$, and $\vy^l \in \sR^{C_{in} \times |G_{in}|}$ where $G_{out}$ and $G_{in}$ are the set of group elements in the output and input spaces respectively. We use the following shorthand for indexing $\vz^l(g) \equiv \vz^{l,:,g} \in \sR^{C_{out}}$ and $\vy^l(g) \equiv \vy^{l, :, g} \in \sR^{C_{in}}$, denoting the vector of feature channels at a specific group element (`fiber'). %(later in Section 3, we extend this to refer to multiple group elements). 
% However, in the following, for the purpose of mathematical analysis (and to match existing notation), we use the standard convention of defining $\vz^l$ and $\vx^l$ as functions over the group (i.e. $\vz^l\ : \ \gG_{out} \rightarrow \mathbb{R}^{C_{out}}$ and $\vx^l\ : \ \gG_{in} \rightarrow \mathbb{R}^{C_{in}}$), and thus denote the feature vectors at each associated group element $g \in \gG$ as $\vz^l(g) \in \sR^{C_{out}}$, and $\vx^l(g) \in \sR^{C_{in}}$ respectively. 
Then, the value $z^{l,c}(g) \in \sR$ of a single output at layer $l$, channel $c$ and element $g$ is 
\begin{equation}
    z^{l,c}(g) \equiv [\vy^l \star \vpsi^{l,c}](g) = \sum_{h \in G_{in}} \sum_{i}^{C_{in}} y^{l, i}(h) \psi^{l, c}_i (g^{-1} \cdot h) \ ,
\end{equation}
where $\vpsi^{l, c}_i$ is the filter between the $i^{th}$ input channel (subscript) and the $c^{th}$ output channel (superscript), and is similarly defined (and indexed) over the set of input group elements $G_{in}$. We highlight that the composition $g^{-1} \cdot h = k \in G_{in}$ is defined by the action of the group and yields another group element by closure of the group. The representation $\Gamma_g$ and can then be defined as $\Gamma_g [\vz^l(h)] = \vz^l(g^{-1} \cdot h)$ for all $l > 0$ when $\gG^l_{in} = \gG^l_{out} = \gG^0_{out}$. From this definition it is straightforward to prove equivariance from: $[\Gamma_g [\vy^l] \star \vpsi^l](h) = \Gamma_g [\vy^l \star \vpsi^l](h) = \Gamma_g[\vz^l](h)$. Furthermore, we see that $\Gamma_g$ is a `regular representation' of the group, meaning that it acts by permuting features along the group dimension while leaving feature channels intact. Group equivariant layers can then be composed with pointwise non-linearities and biases to yield a fully equivariant deep neural network (e.g. $\vy_i^{l+1} = \mathrm{ReLU}(\vz^{l} + \vb)$ where $\vb \in \sR^{C_{out}}$ is a learned bias shared over the output group dimensions). 
For $l=0$, $\vy^0 = \vx$, the raw input, and typically $\gG^0_{in} = (\sZ^2_{HW}, +)$, the group of all 2D integer translations up to height $H$ and width $W$. $\gG^0_{out}$ is then chosen by the practitioner and is typically a larger group which includes translation as a subgroup, e.g. the roto-translation group, or the group of scaling \& translations.

\paragraph{DIM(L) in H-SSL} In this section we outline precisely how the Deep Infomax Local loss DIM(L) relates to the H-SSL framework proposed in Section \ref{sec:s-ssl}. Specifically, in Deep InfoMax (DIM(L)) the same general form of the loss function is applied (often called InfoNCE), but the cosine similarity is replaced with a log-bilinear model: $\mathrm{sim}(\va, \vb) = \mathrm{exp}\left(\va^T W \vb\right)$. Additionally, and most importantly to this work, rather than computing the similarity between two differently augmented versions on an image, the loss is applied between different spatial locations of the representation for a single image, again with a head $h$ applied afterwards. If we let $\vg \sim \sZ_{HW}^2$ refer to sampling a contiguous patch from the spatial coordinates of a convolutional feature map, we can write this general Feature-Space InfoMax loss ($\mathcal{L}_{\text{FSIM}}$) as: 
\begin{equation}
    \footnotesize
    \mathcal{L}_{\text{FSIM}}(\mathbf{X}) = - \frac{1}{N} \sum_{i}^N \E_{\vg_1, \vg_2 \sim \sZ_{HW}^2} \log{\frac{\mathrm{exp}\Big(\mathrm{sim}\big(h(\vz_i(\vg_1)), h(\vz_i(\vg_2))\big) / \tau \Big)}{\sum_{k \neq i}^{N} \sum_{j, l}^2 \mathrm{exp}\Big(\mathrm{sim}\big(h(\vz_i(\vg_j)), h(\vz_k(\vg_l))\big) / \tau\Big)}} \ .
\end{equation}
\normalsize
To show that this is equivalent to our $\mathcal{L}_{\text{H-SSL}}$, we see that the randomly sampled spatial patches $\vg_1, \vg_2$ can equivalently be described as a single base patch $\vg_{0}$ shifted by randomly sampled translations $g_1$ and $g_2$. Explicitly, 
\begin{equation}
    \footnotesize
    \label{eq:fsim}
    \mathcal{L}_{\text{FSIM}}(\mathbf{X}) = - \frac{1}{N} \sum_{i}^N \E_{g_1, g_2 \sim G} \log{\frac{\mathrm{exp}\Big(\mathrm{sim}\big(h(\vz_i(g_1^{-1} \cdot \vg_{0})), h(\vz_i(g_2^{-1} \cdot \vg_{0}))\big) / \tau \Big)}{\sum_{k \neq i}^{N} \sum_{j,l}^2 \mathrm{exp}\Big(\mathrm{sim}\big(h(\vz_i(g_j^{-1} \cdot \vg_{0})), h(\vz_k(g_l^{-1} \cdot  \vg_{0}))\big) / \tau\Big)}} \ .
\end{equation}
\normalsize
Thus, we see that Feature-Space InfoMax losses are included in our framework, and can therefore be seen to be equivalent to input-augmentation based losses with an equivariant backbone, where the set of augmentations is limited to the translation group $G \equiv \sZ_{HW}^2$, and the $\vg_0$ base size is a single spatial coordinate ($|\vg_{0}| = 1$) rather than the size of the full representation ($|\vg_{0}| = |G|$).

%% file: sections/appendix/appendix_E.tex
\section{Broader Impact}
\label{appendix:E}
This work is primarily related to understanding and improving self-supervised learning -- a training method for deep neural networks which is able to leverage large amounts of unlabeled data from the internet, making it one of the most used methods for state of the art image and text generative models today \cite{clip, dalle}. Such models have significant broader impact and potential negative consequences which are beyond the scope of this work. We refer readers to discussions of those paper for further information. Specifically, this work aims to improve such SSL techniques, thereby inheriting the broader impact of these models.